\documentclass[runningheads]{llncs}
\usepackage{makeidx}  % allows for indexgeneration
\usepackage{url}
\usepackage{subcaption}
\usepackage{amsfonts}
\usepackage{amsmath}
\usepackage{color}
\usepackage{hyperref}
\hypersetup{colorlinks=true}

\usepackage{algorithm}
\usepackage{algpseudocode}

\usepackage{xspace}
\newcommand{\etal}{\emph{et al.}\@\xspace}
\newcommand*{\eg}{e.g.\@\xspace}
\newcommand*{\ie}{i.e.\@\xspace}

\usepackage{graphicx}

\begin{document}
\title{Privacy-preserving Federated Brain Tumour Segmentation}

\author{Wenqi Li\inst{1} \and
Fausto Milletar\`\i\inst{1} \and
Daguang Xu\inst{1} \and
Nicola Rieke\inst{1} \and
Jonny Hancox\inst{1} \and
Wentao Zhu\inst{1} \and
Maximilian Baust\inst{1} \and
Yan Cheng\inst{1} \and
S\'{e}bastien Ourselin\inst{2} \and\\
M. Jorge Cardoso\inst{2} \and
Andrew Feng\inst{1}}
\authorrunning{W. Li et al.}
\institute{NVIDIA \and
Biomedical Engineering and Imaging Sciences, King's College London, UK}
\maketitle              % typeset the title of the contribution
\begin{abstract}
Due to medical data privacy regulations, it is often infeasible to collect and
share patient data in a centralised data lake.  This poses challenges for
training machine learning algorithms, such as deep convolutional networks,
which often require large numbers of diverse training examples.  Federated
learning sidesteps this difficulty by bringing code to the patient data owners
and only sharing intermediate model training updates among them. Although a
high-accuracy model could be achieved by appropriately aggregating these model
updates, the model shared could indirectly leak the local training examples.
In this paper, we investigate the feasibility of applying differential-privacy
techniques to protect the patient data in a federated learning setup.  We
implement and evaluate practical federated learning systems for brain tumour
segmentation on the BraTS dataset.  The experimental results show that there is
a trade-off between model performance and privacy protection costs.
\end{abstract}
\section{Introduction}\label{sec:intro}
Deep Neural Networks (DNN) have shown promising results in various medical
applications, but highly depend on the amount and the diversity of training
data~\cite{sun_revisiting_2017}.  In the context of medical imaging,
this is particularly challenging since the required training data
may not be available in a single institution due to the low incidence rate
of some pathologies and limited numbers of patients.
At the same time, it is often infeasible to collect and share
patient data in a centralised data lake due to medical data privacy
regulations.

One recent method that tackles this problem is Federated Learning
(FL)~\cite{mcmahan_communication-efficient_2016,sheller2018multi}: it allows
collaborative and decentralised training of DNNs without sharing the patient
data. Each node trains its own local model and, periodically, submits it to a
parameter server. The server accumulates and aggregates the individual
contributions to yield a global model, which is then shared with all nodes.  It
should be noted that the training data remains private to each node and is
never shared during the learning process. Only the model's trainable weights or
updates are shared, thus keeping patient data private.  Consequently, FL
succinctly sidesteps many of the data security challenges by leaving the data
where they are and enables multi-institutional collaboration.

Although FL can provide a high level of security in terms of privacy, it is
still vulnerable to misuse such as reconstructions of the training examples by
model inversion.  One effective countermeasure is to inject noise to each
node's training process, distort the updates and limit the granularity of
information shared among
them~\cite{abadi_deep_2016,shokri_privacy-preserving_2015}.  However, existing
privacy-preserving research only focuses on general machine learning benchmarks
such as MNIST, and uses vanilla stochastic gradient descent algorithms.

In this work, we implement and evaluate practical federated learning systems
for brain tumour segmentation.  Throughout a series of experiments on the BraTS
2018 data, we demonstrate the feasibility of privacy-preserving techniques.
Our primary contributions are: (1) implement and evaluate, to the best of our
knowledge, the first privacy-preserving federated learning system for medical
image analysis; (2) compare and contrast various aspects of federated averaging
algorithms for handling momentum-based optimisation and imbalanced training
nodes; (3) empirically study the sparse vector technique for a strong
differential privacy guarantee.

\section{Method}\label{sec:method}
\begin{figure}[t]
  \centering
  \includegraphics[width=0.4\linewidth]{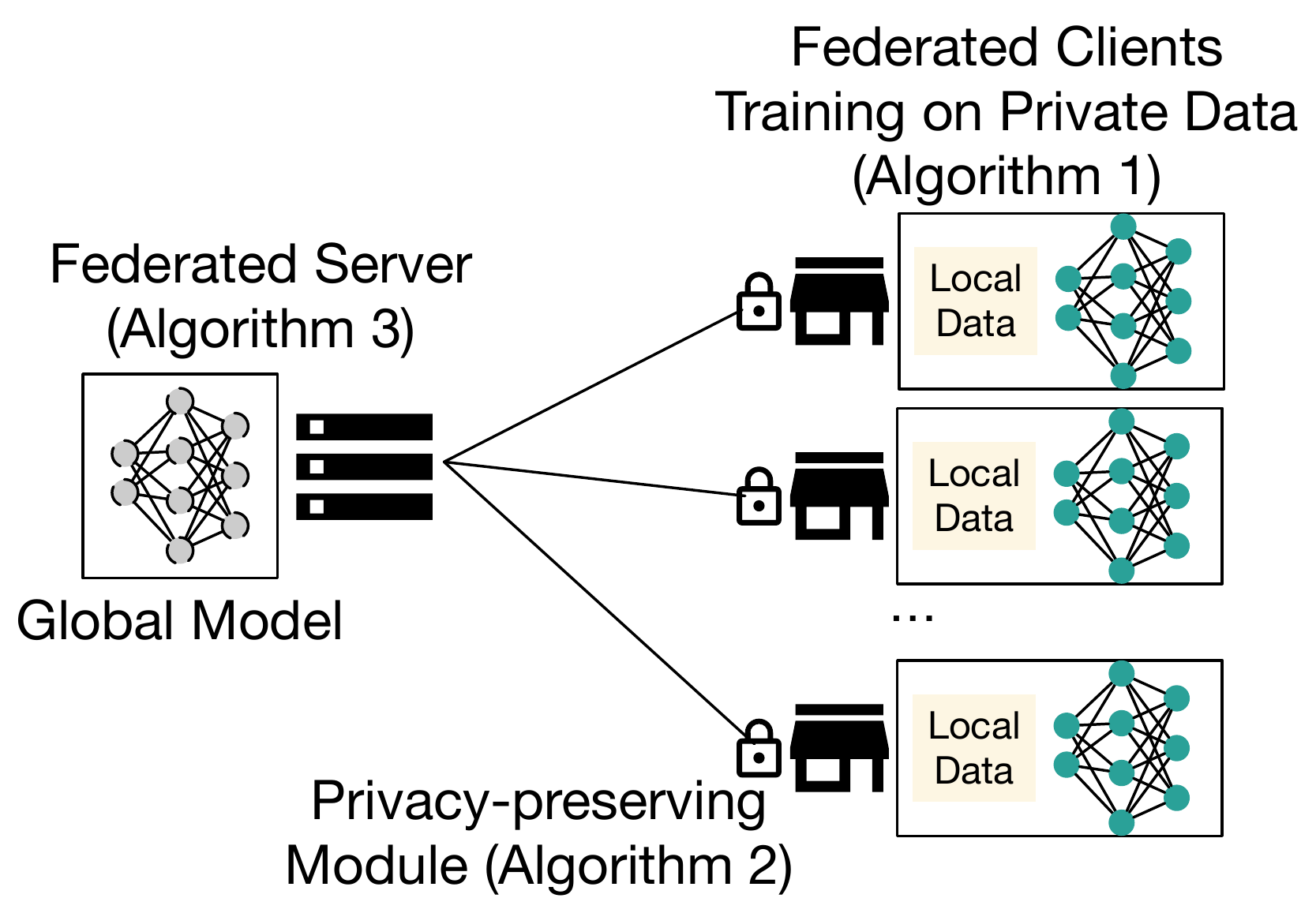}
  \includegraphics[width=0.3\linewidth]{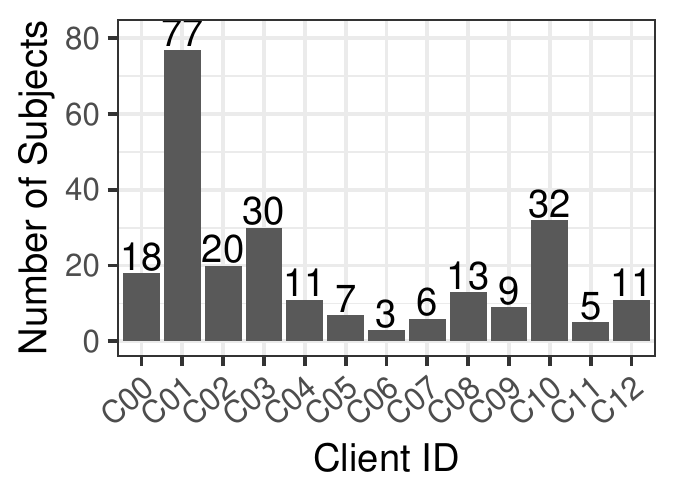}
  \caption{Left: illustration of the federated learning system;
  right: distribution of the training subjects ($N=242$) across the
  participating federated clients ($K=13$) studied in this paper.}\label{fig:FL_system}
\end{figure}

We study FL systems based on a client-server architecture (illustrated in
Fig.~\ref{fig:FL_system} (left)) implementing the federated averaging
algorithm~\cite{mcmahan_communication-efficient_2016}. In this configuration, a
centralised server maintains a global DNN model and coordinates clients' local
stochastic gradient descent (SGD) updates.  This section presents the
client-side model training procedure, the server-side model aggregation
procedure, and the privacy-preserving module deployed on the client-side.

\subsection{Client-side model training}
We assume each federated client has a fixed local dataset and reasonable
computational resources to run mini-batch SGD updates.  The clients also share
the same DNN structure and loss functions.  The proposed training
procedure is listed in Algorithm~\ref{alg:client}.
At federated round $t$, the local
model is initialised by reading global model parameters $W^{(t)}$ from the
server, and is updated to $W^{(l,t)}$ by running multiple iterations of SGD.
After a fixed number of iterations $N^{(local)}$, the model difference
$\Delta W^{(t)}$ is shared with the aggregation server.

DNNs for medical image are often trained with a momentum-based SGD. Introducing
the momentum terms takes the previous SGD steps into account when computing the
current one. It can help accelerate the training and reduce oscillation.  We
explore the choices of design for handling these terms in FL.  In the proposed
Algorithm~\ref{alg:client} (exemplified with Adam
optimiser~\cite{kingma2014adam}), we re-initialise each client's momentums at
the beginning of each federated round (denoted as \texttt{m.\,restart}).  Since
local model parameters are initialised from the global ones, which aggregated
information from other clients, the restarting operation effectively clears the
clients' local states that could interfere the training process.  This is
empirically compared with (a) clients maintaining a set of local momentum
variables without sharing; denoted as \texttt{baseline\ m}. (b) treating the
momentum variables as a part of the model, \ie, the variables are updated
locally and aggregated by the server (denoted as \texttt{m.\,aggregation}).
Although \texttt{m.\,aggregation} is theoretically
plausible~\cite{yu2019linear}, it requires the momentums to be released to the
server. This increases both communication overheads and data security risks.

\begin{algorithm}[t]
  \caption{Federated learning: client-side training at federated round $t$.}\label{alg:client}
  \begin{algorithmic}[1]
    \Require local training data ${\cal{D}} = \{ x_i, y_i \}_{i=1}^{N_c}$, num\_local\_epochs
    \Require learning rate $\eta$, decay rates $\beta_1, \beta_2$, small constant $\epsilon$
    \Require loss function $\ell$ defined on training pairs $(x, y)$ parameterised by $W$
    \Procedure{local\_training}{$\textrm{global model}\ W^{(t)}$}
    \State Set initial local model: $W^{(0, t)} \gets W^{(t)}$
    \State Initialise momentum terms: $m^{(0)} \gets 0$, $v^{(0)} \gets 0$ \label{restart_momentum}
    \State Compute number of local iterations: $N^{(local)}\gets N_c\cdot \textrm{num\_local\_epochs}$
    \For{$l \gets 1\cdots N^{(local)}$}  \textit{\Comment{ Training with Adam optimiser}}
      \State Sample a training batch: ${\cal{B}}^{(l)} \sim {\cal{D}}$
      \State Compute gradient:   $g^{(l)} \gets \nabla\ell({\cal{B}}^{(l)}; W^{(l-1, t)})$
      \State Compute 1st moment: $m^{(l)} \gets \beta_1\cdot m^{(l-1)} + (1 - \beta_1)\cdot g^{(l)}$
      \State Compute 2nd moment: $v^{(l)} \gets \beta_2\cdot v^{(l-1)} + (1 - \beta_2)\cdot g^{(l)}\cdot g^{(l)}$
      \State Compute bias-corrected learning rate: $\eta^{(l)} \gets \eta \cdot \sqrt{1-\beta_2^{l}} / (1 - \beta_1^{l})$
      \State Update local model: $W^{(l, t)} \gets W^{(l-1, t)} - \eta^{(l)} \cdot m^{(l)} / (\sqrt{v^{(l)}} + \epsilon)$
    \EndFor
    \State Compute federated gradient: $\Delta W^{(t)} \gets W^{(l, t)} - W^{(0, t)}$
    \State $\Delta \hat{W}^{(t)} \gets \textrm{PRIVACY\_PRESERVING}(\Delta W^{(t)})$
    \State \Return $\Delta \hat{W}^{(t)}$ and $N^{(local)}$  \textit{\Comment{ Upload to server}}
    \EndProcedure
  \end{algorithmic}
\end{algorithm}

\subsection{Client-side privacy-preserving module}
\begin{algorithm}[t]
  \caption{Federated learning: client-side differential privacy module.}\label{alg:privacy}
  \begin{algorithmic}[1]
    {\small
    \Require privacy budgets for gradient query, threshold, and answer $\varepsilon_1, \varepsilon_2, \varepsilon_3$
    \Require sensitivity $s$, gradient bound and threshold $\gamma, \tau$, proportion to release $Q$
    \Require number of local training iterations $N^{(local)}$
    \Procedure{privacy\_preserving}{$\Delta W$}
      \State Normalise by iterations: $\Delta W \gets \Delta W / N^{(local)}$
      \State Compute number of parameters to share: $q \gets Q \cdot size(\Delta W)$
      \State Track parameters to release: $\Delta \hat{W} \gets empty\ set$
      \State Compute a noisy threshold: $\hat{h} \gets h + Lap(\frac{s}{\varepsilon_2})$
      \While{$size(\Delta \hat{W}) < q$}
        \State Randomly draw a gradient component $w_i$ from $\Delta W$
        \If{$abs(clip(w_i, \gamma)) + Lap(\frac{2qs}{\varepsilon_1}) \geq \hat{h}$}
        \State Compute a noisy answer: $w_i \gets clip(w_i+ Lap(\frac{qs}{\varepsilon_3}), \gamma)$
        \State Release the answer: append $w_i$ to $\Delta \hat{W}$
        \EndIf
      \EndWhile
      \State Undo normalisation: $\Delta \hat{W} \gets \Delta \hat{W} * N^{(local)}$
      \State \Return $\Delta \hat{W}$
    \EndProcedure
  }
  \end{algorithmic}
\end{algorithm}

The client-side is designed to have full control over which data to share and
local training data never leave the client's site.  Still, model inversion
attacks such as~\cite{hitaj2017deep} can potentially extract sensitive patient
data from the update $\Delta W^{(t)}_k$ or the model $W^{(t)}$ during
federated training.  We adopt a selective parameter
update~\cite{shokri_privacy-preserving_2015} and the sparse vector technique
(SVT)~\cite{lyu2017understanding} to provide strong protection against
indirect data leakage.

\textbf{Selective parameter sharing}\label{sec:selective}
The full model at the end of a client-side training process might have
over-fitted and memorised local training examples.  Sharing this model poses
risks of revealing the training data.  Selective parameter sharing methods
limit the amount of information that a client shares.  This is achieved by (1)
only uploading a fraction of $\Delta W^{(t)}_k$: component $w_i$ of $\Delta
W^{(t)}_k$ will be shared iif $abs(w_i)$ is greater than a threshold
$\tau^{(t)}_k$; (2) further replacing $\Delta W^{(t)}_k$ by clipping the values
to a fixed range $[-\gamma, \gamma]$.  Here $abs(x)$ denotes the absolute value
of $x$; $\tau^{(t)}_k$ is chosen by computing the percentile of $abs(\Delta
W^{(t)}_k)$; $\gamma$ is independent of specific training data and can be
chosen via a small publicly available validation set before training.  Gradient
clipping is also applied, which is a widely-used method,
acting as a model regulariser to prevent over-fitting.

\textbf{Differential privacy module}\label{sec:dp}
The selective parameter sharing can be further improved by having a strong
differential privacy guarantee using SVT.  The procedure of selecting and
sharing distorted components of $w_i$ is described in
Algorithm~\ref{alg:privacy}.  Intuitively, instead of simply thresholding
$abs(\Delta W^{(t)}_k)$ and sharing its components $w_i$, every
sharing $w_i$ is controlled by the Laplacian mechanism. This is implemented by
first comparing a clipped and noisy version of $abs(w_i)$ with a noisy
threshold $\tau^{(t)} + Lap(s/\varepsilon_2)$ (Line 8,
Algorithm~\ref{alg:privacy}), and then only sharing a noisy answer $clip(w_i +
Lap(qs/\varepsilon_3), \gamma)$, if the thresholding condition is satisfied.
Here $Lap(x)$ denotes a random variable sampled from the Laplace distribution
parameterised by $x$; $clip(x, \gamma)$ denotes clipping of $x$ to be in the
range of $[-\gamma, \gamma]$; $s$ denotes the sensitivity of the federated
gradient which is bounded by $\gamma$ in this
case~\cite{shokri_privacy-preserving_2015}.  The selection procedure is
repeated until $q$ fraction of $\Delta W^{(t)}_k$ is released.  This procedure
satisfies $(\varepsilon_1+\varepsilon_2+\varepsilon_3)$-differential
privacy~\cite{lyu2017understanding}.

\subsection{Server-side model aggregation}
The server distributes a global model and receives synchronised updates from
all clients at each federated round (Algorithm~\ref{alg:server}).  Different
clients may have different numbers of local iterations at round $t$, thus the
contributions from the clients could be SGD updates at different training
speeds.  It is important to require an $N^{(local)}$ from the clients, and
weight the contributions when aggregating them (Line 8,
Algorithm~\ref{alg:server}).  In the case of partial model sharing, utilising
the sparse property of $\Delta W^{(t)}_k$ to reduce the communication overheads
is left for future work.

\begin{algorithm}[t]
  \caption{Federated learning: server-side aggregation of $T$ rounds.}\label{alg:server}
  \begin{algorithmic}[1]
    {\small
    \Require{num\_federated\_rounds}
    \Procedure{Aggregating}{}
      \State{Initialise global model: $W^{(0)}$}
      \For{$t \gets 1\cdots T$}
        \For{$client\ k \gets 1\cdots K$}  \Comment{ \textit{Run in parallel}}
        \State{Send $W^{(t-1)}$ to client $k$}
        \State{Receive $(\Delta W_k^{(t-1)}, N_k^{(local)})$ from client's $\textrm{LOCAL\_TRAINING}(W^{(t-1)})$}
        \EndFor
        \State{$W^{(t)}\gets W^{(t-1)} + \frac{1}{\sum_k{N_k^{(local)}}}\sum_k{(N_k^{(local)}\cdot \Delta W_k^{(t-1)})}$}
      \EndFor
      \State \Return $W^{(t)}$
    \EndProcedure
    }
  \end{algorithmic}
\end{algorithm}

\section{Experiments}\label{sec:experiments}
This section describes the experimental setup, including the common
hyper-parameters used for each FL system.

\textbf{Data preparation}
The BraTS 2018 dataset~\cite{bakas2018identifying} contains multi-parametric
pre-operative MRI scans of 285 subjects with brain tumours.  Each subject was
scanned with four modalities, \ie(1) T1-weighted, (2) T1-weighted with contrast
enhancement, (3) T2-weighted, and (4) T2 fluid-attenuated inversion recovery
(T2-FLAIR).  Each subject was associated with voxel-level annotations of
``whole tumour'', ``tumour core'', and ``enhancing tumour''.  For details of
the imaging and annotation protocols, we refer the readers to Bakas
\etal~\cite{bakas2018identifying}.  The dataset was previously used for
benchmarking machine learning algorithms and is publicly available.  We use it
to evaluate the FL algorithms on the multi-modal and multi-class segmentation
task.  For the client-side local training, we adapted the state-of-the-art
training pipeline originally designed for data-centralised
training~\cite{myronenko20183d} and implemented as a part of the NVIDIA Clara
Train SDK\footnote{\scriptsize https://devblogs.nvidia.com/annotate-adapt-model-medical-imaging-clara-train-sdk/}.

To test the generalisation ability across the subjects, we randomly split the
dataset into a model training set ($N=242$ subjects) and a held-out test set
($N=43$ subjects).  The scans were collected from thirteen institutions with
different equipment and imaging protocols, and thus heterogeneous
image feature distributions.  To make our federated setup realistic, we further
stratified the training set into thirteen disjoint subsets, according to where
the image data were originated and assigned each to a federated client.  The
setup is challenging for FL algorithms, because (1) each client only
processes data from a single institution, which potentially suffers from more
severe domain-shift and over-fitting issues compared with a data-centralised
training; (2) it reflects the highly imbalanced nature of the dataset (shown in
Fig.~\ref{fig:FL_system}).

\textbf{Federated model setup}
The evaluation of the FL procedures is perpendicular to the choice of
convolutional network architectures.  Without loss of generality, we chose the
segmentation backbone of~\cite{myronenko20183d} as the underlying federated
model and used the same set of local training hyperparameters for all
experiments: the input image window size of the network was
$224\times224\times128$ voxels, and spatial dropout ratio of the first
convolutional layer was $0.2$.  Similarly to~\cite{myronenko20183d}, we
minimised a soft Dice loss using Adam~\cite{kingma2014adam} with a learning
rate of $10^{-4}$, batch size of $1$, $\beta_1$ of $0.9$, $\beta_2$ of $0.999$,
and $\ell_2$ weight decay coefficient of $10^{-5}$.  For all federated
training, we set the number of federated rounds to $300$ with two local epochs
per federated round.  A local epoch is defined as every client ``sees'' its
local training examples exactly once.  At the beginning of each epoch, data
were shuffled locally for each client.  For a comparison of model convergences,
we also train a data-centralised baseline for $600$ epochs.

In terms of computational costs, the segmentation
model has about $1.2\times10^6$ parameters; a training iteration with
an NVIDIA Tesla V100 GPU took $0.85\,\textrm{s}$.

\textbf{Evaluation metrics}
We measure the segmentation performance of the models on the held-out test set
using mean-class Dice score averaged over the three types of tumour regions and
all testing subjects.  For the FL systems, we report the performance of the
global model shared among the federated clients.

\textbf{Privacy-preserving setup}
The selective parameter updates module has two system parameters: fraction of the
model $q$ and the gradient clipping value $\gamma$.  We report model
performance by varying both.
For differential privacy, we fixed $\gamma$ to $10^{-4}$, the
sensitivity $s$ to $2\gamma$, and $\varepsilon_2$ to
$(2qs)^{\frac{2}{3}}\varepsilon_1$ according to~\cite{lyu2017understanding}.
The model performance by varying $q$, $\varepsilon_1$,
and $\varepsilon_3$ are reported in the next section.

\section{Results}
\textbf{Federated vs.\ data-centralised training}
\begin{figure}[t]
  \centering
  \includegraphics[width=0.38\linewidth]{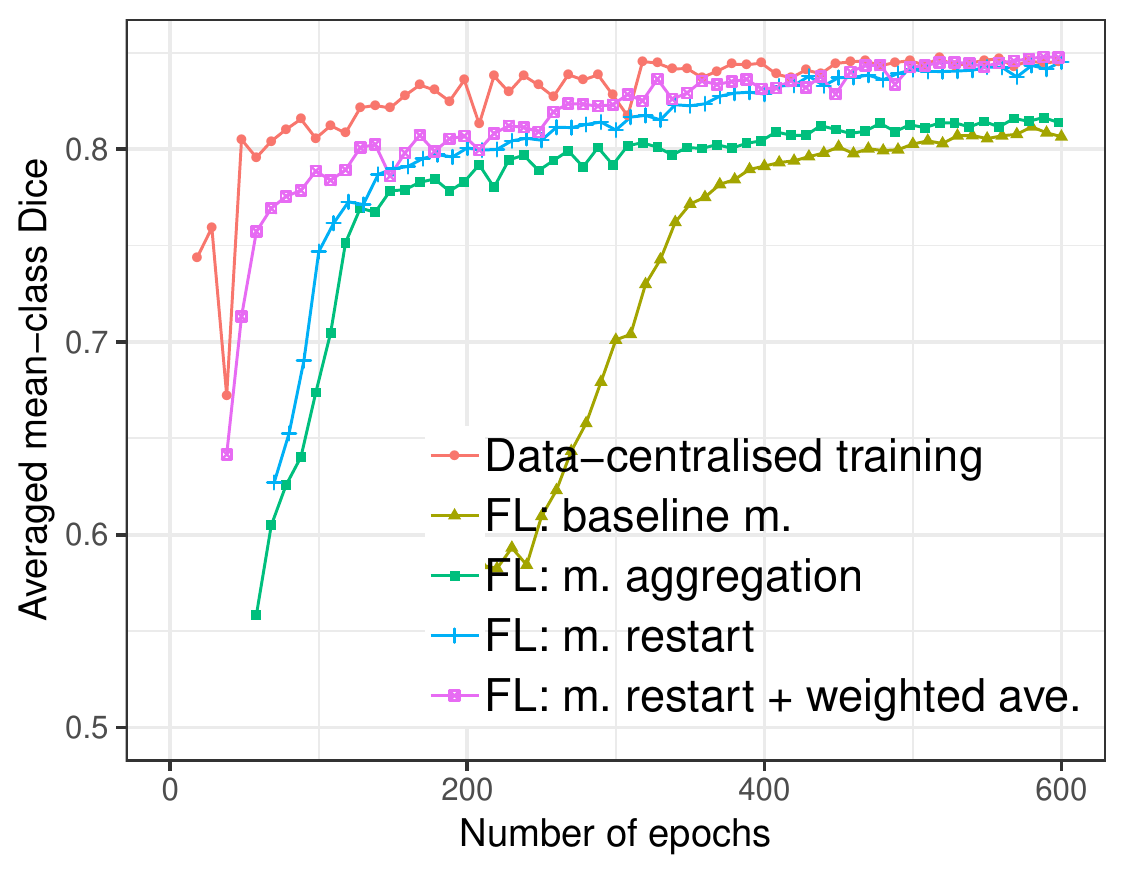}
  \includegraphics[width=0.38\linewidth]{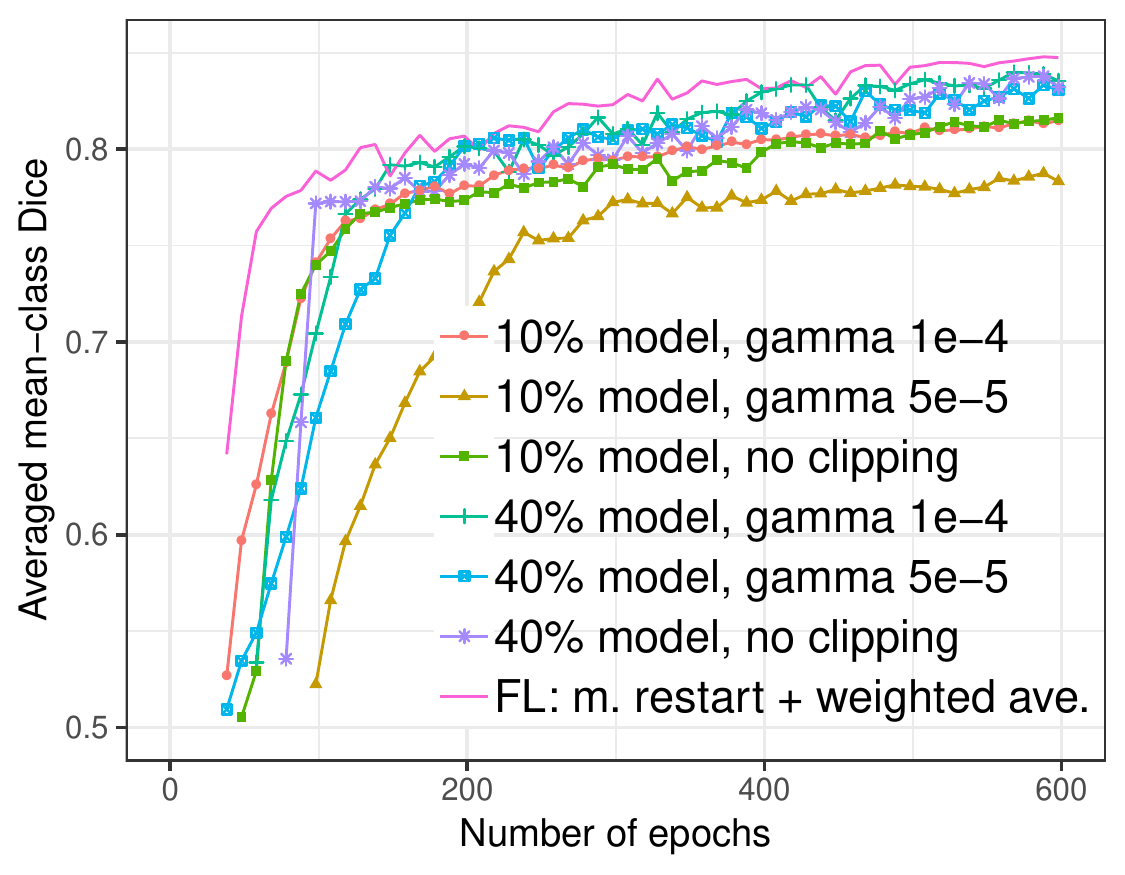}
  \caption{Comparison of segmentation performance on the test set with
  (left): FL vs.\ non-FL training, and (right): partial model sharing.}
  \label{fig:fl_non_fl}
\end{figure}
The FL systems are compared with the data-centralised training in
Fig.~\ref{fig:fl_non_fl} (left).  The proposed FL procedure can achieve a
comparable segmentation performance without sharing clients' data.  In terms of
training time, the data-centralised model converged at about $300$ training
epochs, FL training at about $600$.  In our experiments, an epoch of
data-centralised training ($N=242$) with an NVIDIA Tesla V100 GPU takes $0.85s
\times 242 = 205.70s$ per epoch. The FL training time was determined by the
slowest client ($N=77$), which takes $0.85s \times 77 = 65.45s$ plus small
overheads for client-server communication.

\textbf{Momentum restarting and weighted averaging}
Fig.~\ref{fig:fl_non_fl} (left) also compares variants of the FL procedure.
For the treatment of momentum variables, restarting them at each federated
round outperforms all the other variants.  This suggests (1) each client
maintaining an independent set of momentum variables slows down the convergence
of the federated model;  (2) averaging the momentum variables across clients
improved the convergence speed over \texttt{baseline\,m.}, but still gave a
worse global model than the data-centralised model.  On the server-side,
weighted averaging of the model parameters outperforms the simple model
averaging (\ie $W^{(t+1)} \gets {\sum_k W^{(t+1)}_k}/{K}$).  This suggests that
the weighted version can handle imbalanced numbers of iterations across the
clients.

\textbf{Partial model sharing}
Fig.~\ref{fig:fl_non_fl} (right) compares partial model sharing by varying the
fraction of the model to share and the gradient clipping values.  The figure
suggests that sharing larger proportions of models can achieve better
performance. Partial model sharing does not affect the model convergence speed
and the performance decrease can be almost negligible when only 40\% of the
full model is shared among the clients. Clipping of the gradient can, sometimes,
improve the model performance. However, the value needs to be carefully tuned.

\begin{figure}[t]
  \centering
  \begin{subfigure}{0.24\textwidth}
    \includegraphics[width=\textwidth]{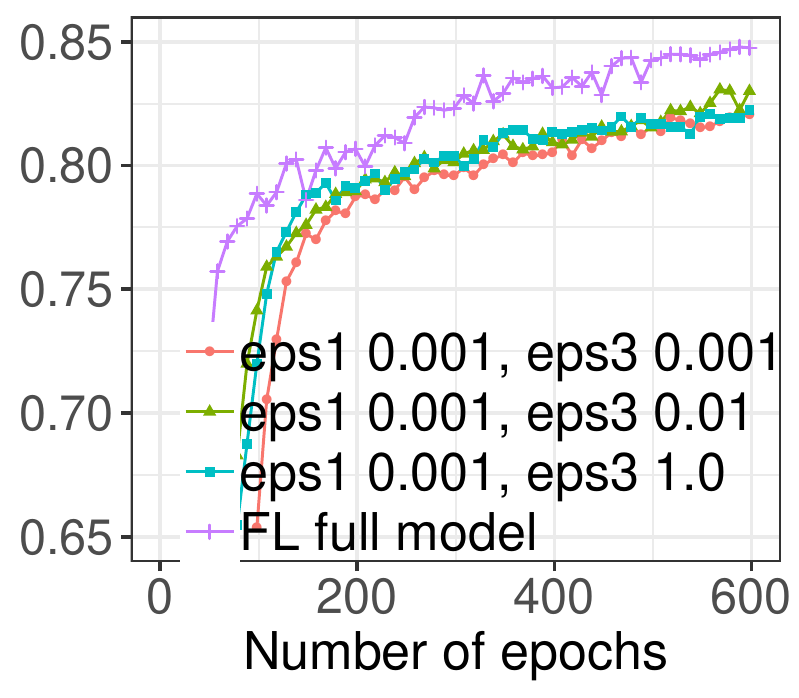}
    \caption{10\% model sharing}
  \end{subfigure}
  \begin{subfigure}{0.24\textwidth}
    \includegraphics[width=\textwidth]{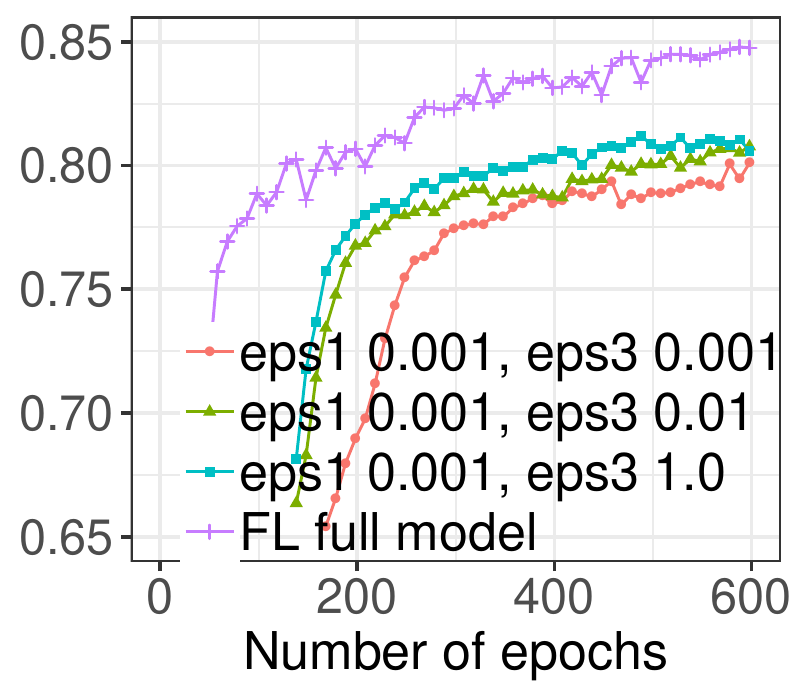}
    \caption{40\% model sharing}
  \end{subfigure}
  \begin{subfigure}{0.24\textwidth}
    \includegraphics[width=\textwidth]{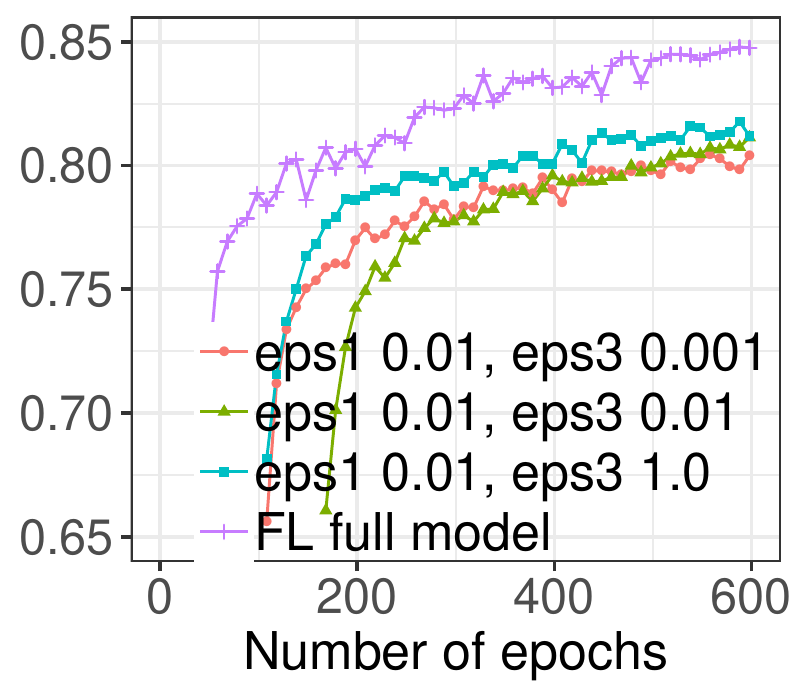}
    \caption{40\% model sharing}
  \end{subfigure}
  \begin{subfigure}{0.24\textwidth}
    \includegraphics[width=\textwidth]{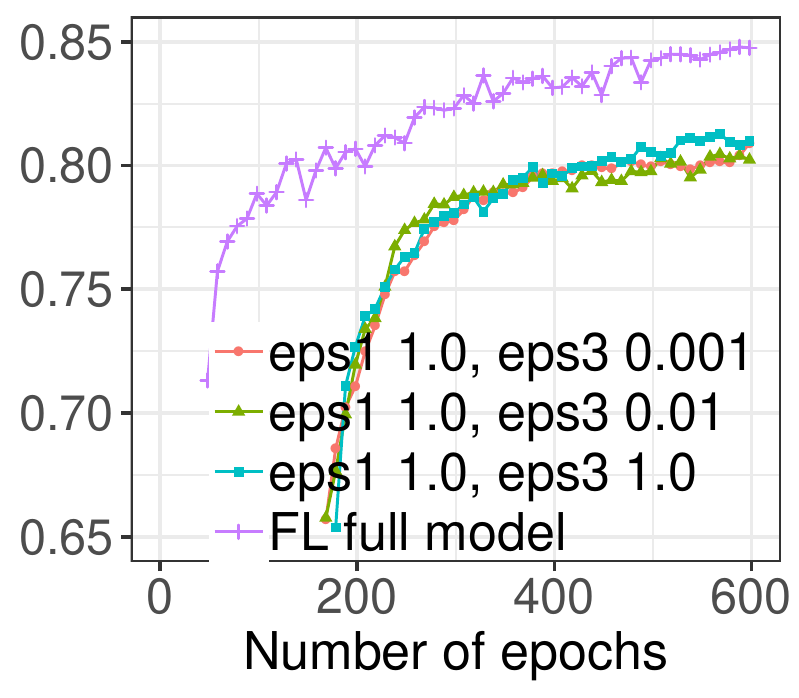}
    \caption{40\% model sharing}
  \end{subfigure}
  \caption{Comparison of segmentation models (ave.\ mean-class Dice
  score) by varying the privacy parameters: percentage of
  partial models, $\varepsilon_1$, and $\varepsilon_3$.}\label{fig:fl_dp}
\end{figure}

\textbf{Differential privacy module}
The model performances by varying differential privacy (DP) parameters are
shown in Fig.~\ref{fig:fl_dp}. As expected, there is a trade-off between DP
protection and model performance.  Sharing 10\% model showed better performance
than sharing 40\% under the same DP setup.  This is due to the fact that the
overall privacy costs $\varepsilon$ are jointly defined by the amount of noise
added and the number of parameters shared during training.  By fixing the
per-parameter DP costs, sharing fewer variables has less overall DP costs and
thus better model performance.

\vspace{-0.05cm}
\section{Conclusion}
\vspace{-0.05cm}
We propose a federated learning system for brain tumour segmentation. We
studied various practical aspects of the federated model sharing with an
emphasis on preserving patient data privacy.  While a strong
differential privacy guarantee is provided, the privacy cost allocation is
conservative. In the future, we will explore differentially private SGD
(\eg~\cite{abadi_deep_2016}) for medical image analysis tasks.

\medskip
\noindent\textbf{Acknowledgements:}
\small
We thank Rong Ou at NVIDIA for the helpful discussions.
\\The research was supported by
the Wellcome/EPSRC Centre for Medical Engineering (WT203148/Z/16/Z),
the Wellcome Flagship Programme (WT213038/Z/18/Z),
the UKRI funded London Medical Imaging and AI centre for
Value-based Healthcare, and the NIHR Biomedical Research Centre
based at Guy's and St Thomas' NHS Foundation Trust and King's College London.
The views expressed are those of the authors and not necessarily those of the
NHS, the NIHR or the Department of Health.

\bibliographystyle{splncs03}
\small \bibliography{references}
\end{document}